# On Learning by Exchanging Advice

Luís Nunes
Eugénio Oliveira
LIACC-NIAD&R;
FEUP Av. Dr. Roberto Frias 4200-465, Porto, Portugal.
Luis.Nunes@iscte.pt; eco@fe.up.pt

**Abstract**

One of the main questions concerning learning in Multi-Agent Systems is: "(How) can agents benefit from mutual interaction during the learning process?". This paper describes the study of an interactive advice-exchange mechanism as a possible way to improve agents' learning performance. The advice-exchange technique, discussed here, uses supervised learning (backpropagation), where reinforcement is not directly coming from the environment but is based on advice given by peers with better performance score (higher confidence), to enhance the performance of a heterogeneous group of Learning Agents (LAs). The LAs are facing similar problems, in an environment where only reinforcement information is available. Each LA applies a different, well known, learning technique: Random Walk (hill-climbing), Simulated Annealing, Evolutionary Algorithms and Q-Learning. The problem used for evaluation is a simplified traffic-control simulation. In the following text the reader can find a description of the traffic simulation and Learning Agents (focused on the advice-exchange mechanism), a discussion of the first results obtained and suggested techniques to overcome the problems that have been observed. Initial results indicate that advice-exchange can improve learning speed, although "bad advice" and/or blind reliance can disturb the learning performance. The use of supervised learning to incorporate advice given from non-expert peers using different learning algorithms, in problems where no supervision information is available, is, to the best of the authors' knowledge, a new concept in the area of Multi-Agent Systems Learning.

## 1 Introduction

### 1.1 Framework

The objective of this work is to contribute to an answer to the question: "(How) can agents benefit from mutual interaction during the learning process, in order to achieve better individual and overall system performances?". This question has been deemed a "challenging issue" by several authors in recently published work (Sen, 1996; Weiß and Dillenbourgh, 1999; Kazakov and Kudenko, 2001; Matarić, 2001).

In the pursuit of an answer to this question, the objects of study are the interactions between the Learning Agents (hereafter referred as agents for the sake of simplicity) and the effects these interactions have on individual and global learning processes. Interactions that affect the learning process can take several forms in Multi-Agent Systems (MAS). These forms range from the indirect effects of other agents' actions (whether they are cooperative or competitive), to direct communication of complex knowledge structures, as well as cooperative negotiation of a search policy or solution.

The most promising way in which cooperative learning agents can benefit from interaction seems to be by exchanging (or sharing) information regarding the learning process itself. As observed by Tan (1993) agents can exchange information regarding several aspects of the learning process: a) the state of the environment, b) episodes (state, action, reward triplets), or c) internal parameters and policies.

Exchanging environment states can be seen as a form of shared exploration. Sharing this information may require a large amount of communication, although the use of a selective policy for the exchange of information may reduce this cost. This type of interaction may be seen as if each agent has extra sets of sensors spread out in the environment, being able to have a more complete view of its external state. This larger view of the state space may require either pre-acquired knowledge on how to interpret this information and integrate it with its own view of the environment's state, or simply be considered as extra input providing a wider range of information about the state. In the limit case, where all agents have access to information regarding the state sensed by all their peers, each agent could be seen as a classical Machine Learning (ML) system with distributed sensors if we consider other agents' actions as part of the environment. One interesting difference, though, is the fact that other agents sensors are not under the control of the learning agent and the perspective they provide on the world may be biased by the needs of the owner of the sensor.

Episode exchange requires that the agents are (or have been) facing similar problems, requiring similar solutions and may also lead to large amounts of communication if there is no criteria regulating the exchange of information. In the limit case, where all agents share all the episodes, this process can also be seen as a single learning system, and produce very little new knowledge. In fact, the exchange of too much data could lead all the agents to follow the same path through the search space, wasting valuable exploration resources.

Sharing internal parameters is another way in which agents can benefit from the knowledge obtained by their peers. Again, in the limit, this can be seen as the use of a single learning agent if communication is unrestricted. This type of information exchange requires that agents have similar internal structures, so that they can easily map their peers' internal parameters into their own, or that they share a complex domain ontology.

As can be seen in the above paragraphs the question is not only: "what type of information to exchange?", but also "when to exchange information?" and "how much information is it convenient to exchange?". When considering human cooperative learning in a team, a common method to improve one's skills is to ask for advice at critical times, or request a demonstration of a solution to a particular problem to someone who is reputed to have better skills in the subject. This is what we have attempted to translate into the realm of Multi-Agent Systems Learning (MASL). Another interesting outcome of our experiments concerns the degree of adequacy different algorithms, being used by the agents, exhibit for different situations considered in the scenario. However, this is not of great importance where MASL is concerned, except in which regards an agent's knowledge about whom to request an advice to in each particular situation. It is our hope that different agents specialise in different situations, becoming, up to a certain extent, complementary in the Multi-Agent System context.

## 1.2 Rationale and summarized description

This paper reports experiments in which agents selectively share episodes by requesting advice for given situations to other agents whose score is, currently, better than their own in solving a particular problem. Considering the discussion of the previous section, this option seemed the most promising for the following reasons:

a) Sharing of episodes does not put heavy restrictions on the heterogeneity of the underlying learning algorithms;

b) Having different algorithms solving similar problems may lead to different forms of exploration of the same search space, thus increasing the probability of finding a good solution;

c) It is more informative and less dependent on pre-coded knowledge than the exchange of environment's states.

Experiments were conducted with a group of Learning Agents embedded in a simplified simulation of a traffic control problem to test the advantages and problems of advice-exchange during learning. Each individual agent uses a standard version of a well know, sub-symbolic, learning algorithm (Random Walk, Evolutionary Algorithms, Simulated Annealing, and Q-Learning). Agents are heterogeneous (i.e., each applies a different learning mechanism, unknown to others). This fact makes communication of internal parameters or policies suffer from the above-mentioned disadvantages, thus it was not considered. The information exchanged amongst agents is: current state (as seen by the advisee agent); best response that can be provided to that state (by the advisor agent); present and best scores, broadcasted at the end of each training stage (epoch).

The problem chosen to test the use of advice-exchange has, as most problems studied in MASL, the following characteristics:

a) Analytical computation of the optimal actions is intractable;

b) The only information available to evaluate learning is a measure of the quality of the present state of the system;

c) The same action executed by a given agent may have different consequences at different times, even if the system is (as far as the agent is allowed to know) in the same state;

d) The agent has only a partial view of the problem's state.

The simplified traffic control problem chosen for these experiments requires that each agent learn to control the traffic-lights in one intersection under variable traffic conditions. Each intersection has four incoming, and four outgoing, lanes. One agent controls the four traffic lights necessary to discipline traffic in one intersection. In the experiments reported here, the crossings controlled by each of the agents are not connected.

The learning parameters of each agent are adapted using two different methods: a reinforcement-based algorithm, using a quality measure that is directly supplied by the environment, and supervised learning using the advice given by peers as the desired response. Notice that the term "reinforcement-based" is used to mean "based on a scalar quality/utility feedback", as opposed to supervised learning which requires a desired response as feedback. The common usage of the term "reinforcement learning", that refers to variations of temporal difference methods (Sutton and Barto, 1987), is a subclass of reinforcement-based algorithms, as are, for instance, most flavours of Evolutionary Algorithms.

## 2 Related Work

The advantages and drawbacks of sharing information and using external teachers in variants of Q-Learning (Watkins and Dayan, 1992) had some important contributions in the early 90's. Whitehead (1991) reports on the usage of two cooperative learning mechanisms: Learning with an External Critic (LEC) and Learning By Watching (LBW). The first, (LEC), is based on the use of an external automated critic, while the second (LBW), learns vicariously by watching other agent's behaviour (which is equivalent to sharing state, action, quality triplets). This work proves that the complexity of the search mechanisms of both LEC and LBW is inferior to that of standard Q-Learning for an important class of state-spaces. Experiments reported in (Whitehead and Ballard, 1991) support these conclusions.

Lin (1992) uses a human teacher to improve the performance of two variants of Q-Learning. This work reports that the "advantages of teaching should become more relevant as the learning task gets more difficult". Results presented show that teaching does improve learning performance in the harder task tested (a variant of the maze problem), although it seems to have no effect on the performance on the easier task (an easier variant of the same maze problem).

The main reference on related work is (Tan, 1993). Tan addressed the problem of exchanging information during the learning process amongst Q-Learning agents. This work reports the results of sharing several types of information amongst several (Q-Learning) agents in the predator-prey problem. Experiments were conducted in which agents shared policies, episodes (state, action, quality triplets), and sensation (state). Although the experiments use solely Q-Learning in the predator-prey domain, the author believes that: "conclusions can be applied to cooperation among autonomous learning agents in general". Conclusions point out that "a) additional sensation from another agent is beneficial if it can be used efficiently, b) sharing learned policies or episodes among agents speeds up learning at the cost of communication, and c) for joint tasks, agents engaging in partnership can significantly outperform independent agents, although they may learn slowly in the beginning". Results presented in (Tan, 1993) also appear to point to the conclusion that sharing episodes with peers is beneficial and can lead to a performance similar to that obtained by sharing policies. Sharing episodes volunteered by an expert agent leads to the best scores in the presented tests, significantly outperforming all other agents in the experiments.

After these first, fundamental, works several variants of information sharing Q-Learners appeared reporting good results in the mixture of some form of teaching and reinforcement learning.

Baroglio (1995) uses an automatic teacher and a technique called "shaping" to teach a Reinforcement Learning algorithm the task of pole balancing. Shaping is defined as a relaxation of the evaluation of goal states in the beginning of training, and a tightening of those conditions in the end.

Clouse (1996) also uses an automatic expert trainer to give the agent actions to perform, thus reducing the exploration time.

Matarić (1996) reports on the use of localized communication to share sensory data and reward as a way to overcome hidden state and credit assignment problems in groups of agents. The experiments conducted in two robot problems, (block pushing and foraging) show improvements in performance on both cases. Later work by the same author, (Matarić, 2001) reports several good results using human teaching and learning by imitation in robot tasks. Experimental results can be found in (Jenkins et al. 2000; Nicolescu and Matarić, 2001; Matarić, 2001b).

Brafman and Tenemholtz (1996) use an expert agent to teach a student agent in a version of the "prisoner's dilemma". The agents implement variations of Q-Learning.

Maclin and Shavlik (1997) use human advice, encoded in rules, which are acquired in a programming language that was specially designed for this purpose. These rules are inserted in a Knowledge Based Neural Network (KBANN) used in Q-Learning to estimate the quality of a given action.

Berenji and Vengerov (2000) report analytical and experimental results concerning the cooperation of Q-Learning agents by sharing quality values amongst them. Experiments were conducted in two abstract problems. Results point out that limitations to cooperative learning described in (Whitehead, 1991) can be surpassed successfully under certain circumstances, leading to better results than the theoretical predictions foresaw.

Simultaneous uses of Evolutionary Algorithms (Holland, 1975; Koza, 1992) and Backpropagation (Rumelhart, Hinton and Williams 1986) are relatively common in Machine Learning (ML) literature, although in most cases Evolutionary Algorithms are used to select the topology or learning parameters, and not to update weights. Some examples can be found in (Salustowicz, 1995) and (Yao, 1999). There are also reports on the successful use of Evolutionary Algorithms and Backpropagation simultaneously for weight adaptation (Topchy, Lebedko and Miagkikh, 1996; Ku and Mak, 1997; Ehardh et al. 1998). Most of the problems in which a mixture of Evolutionary Algorithms and Backpropagation is used are supervised learning problems, i.e., problems for which the desired response of the system is known in advance (not the case of the problem studied in this paper). Castillo et al. (1998) obtained good results in several standard ML problems using Simulated Annealing and Backpropagation, in a similar way to that which is applied in this work. Again, this was used as an add-on to supervised learning to solve a problem for which there is a well known desired response.

The use of learning techniques for the control of traffic-lights can be found in (Goldman and Rosenschein, 1995; Thorpe, 1997; Brockfeld et al. 2001).

## 3 Experimental Setup

This section will describe the internal details of the traffic simulation, the learning mechanisms and the advice-exchange technique.

### 3.1 The Traffic Simulator

The traffic simulator environment is composed of lanes, lane-segments, traffic-lights (and the corresponding con-

trolling agents), and cars. Cars are "well behaved", in the sense that they:

a) Can only move forward;
b) Do not cross yellow or red-lights;
c) Move at a constant speed;
d) Do not crash into other cars.

Cars are inserted at the beginning of each lane, whenever that space is empty, with a probability that varies in time according to a saw-tooth function, of the form:

$$pInsert(t) = (M - m)((t + T_0)\%T)/T + m \quad (1)$$

where $T_0$ is the initial delay, $T$ is the period, $m$ the minimum probability for car insertion and $M$ the maximum. The parameters used in the experiments discussed here for the generation of new cars in the beginning of each lane were in the following ranges: $0 < T_0 \leq T$, $50 \leq T \leq 5000$, $0.01 \leq m \leq 0.1$, $0.01 \leq M \leq 0.3$. In further experiments different generation functions were used, mostly based in superimposition of gaussian functions, but the results reported here were acquired using the saw-tooth generation-function.

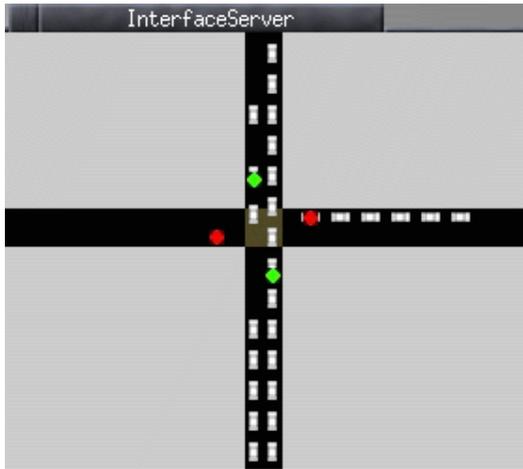

Figure 1: A screenshot of the graphic interface for the Traffic Simulator, showing a partial view of a local scenario.

The time-unit used throughout this description is one turn. One turn corresponds to a period where each object in the system is allowed to perform one action and all the necessary calculations for it. Lanes have three lane-segments: incoming (before the crossing, where cars are inserted), crossing and outgoing. Each local scenario (Figure 1) consists of four lanes, each with a different movement direction and one crossing (the lanes in a local scenario will be referred as North, South, East and West, for the remainder of this description). In the experiments reported here the local scenarios are not connected, i.e., each lane has only one crossing and one traffic light. Cars are inserted in its incoming lane-segment and removed when they reach the extremity of its outgoing lane-segment, after having passed the crossing. Each incoming lane-segment was designed to hold a maximum of 60 cars.

At the beginning of each green-yellow-red cycle, the agents observe the state of environment for their local scenario and decide on the percentage of green-time ($g$) to attribute to the North and South lanes (the percentage of time attributed to the East and West lanes is automatically set at $1 - g$. Yellow-time is fixed in each experiment and lies in the interval [10, 15] turns). Two types of description of the environment's state are used, the first is realistic in the sense that it is technically achievable to collect that type of data in a real situation and it is actually used by traffic controllers today. The second, although it may be unfeasible in today's traffic monitoring systems, was considered to have relevant information for the learning process.

In the first type of state representation, the state $s(t)$, at time $t$, is composed by four scalar values ($s_N$, $s_S$, $s_E$, $s_W$), where each component ($s_i$) represents the ratio of the number of incoming vehicles ($n_i(t)$) in lane $i$ relative to the total number of incoming vehicles in all lanes. This state representation will be referred as *count state representation*.

$$s_i(t) = \frac{n_i(t)}{\sum_j n_j(t)} \quad (i, j \in \{N, S, E, W\}) \quad (2)$$

The second type of environment state has the same information as the one described above plus four scalar values, each of which represents the lifetime (number of turns since creation) of the incoming vehicle that is closest to the traffic-light ($life\_first_i(t)$). To keep inputs within the interval [0,1], this value was cut-off at a maximum lifetime ($lifemax$), and divided by the same value. Thus, the four extra scalar values are:

$$s_i(t) = \frac{life\_first_i(t)}{life\max} \quad (i, j \in \{N, S, E, W\}) \quad (3)$$

if $life\_first(t) < lifemax$ or 1 otherwise. The value of $lifemax$ was chosen to be 3 to 10 times the number of turns a car takes to reach the crossing at average speed, depending on the difficulty of each particular scenario, which is mainly dependent on the parameters used for car generation. This state representation will be referred as *count-time state representation*. The state representations described above are similar to the ones that were reported to have produced some of the best results in the experiments conducted by Thorpe (1997) for the same type of problem (learning to control traffic-lights at an intersection). The normalization of the inputs to fit the [0,1] interval was necessary, even at the cost of loss of information, for two main reasons: a) it keeps the first layer of sigmoids from reaching saturation too early in the learning process; b) using percentages for the first four elements of the state space allows a substantial reduction of the number of possible states, as described below when the implementation of Q-Learning is discussed.

The quality of service of each traffic-light controller at time $t$, is given by $q(t)$, which was initially calculated according to

$$q(t) = 1 - \frac{\sum_i life_i(t)/n}{life\max}, \quad (4)$$

where $life_i(t)$ is the number of turns since creation of car $i$ at time $t$ and $lifemax$ has the same meaning as above. The sum is made for all ($n$) cars in the incoming lane-segments of a crossing. This measure did not provide enough differentiation of "good" and "bad" environment states, thus a logistic function was introduced, using $q(t)$, in (4), as input, to emphasize the difference in quality between these two types of environment states. A comparative view of both functions can be seen in Figure 2, the former in continuous line the latter in dashed line style.

The car generation parameters in traffic simulator proved difficult to tune. Slight changes led to simulations that were either too difficult (no heuristic nor any learned strategy were able to prevent major traffic jams), or to problems in which both simple heuristics and learned strategies were able to keep a normal traffic flow with very few learning steps.

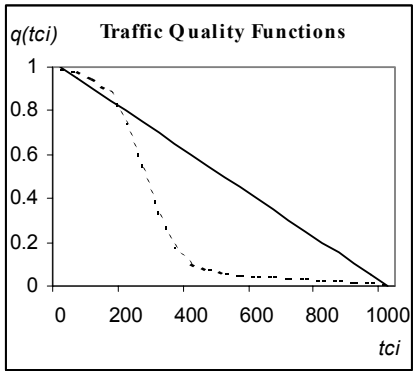

Figure 2: Two functions for the evaluation of traffic quality based on the average time of life of the incoming cars (*tci*).

The traffic simulator was coded in C++, with a Java graphical interface (Figure 1). Agents are not independent processes, at this stage they are merely C++ objects that are given a turn to execute their actions in round-robin. On the one hand, this choice eliminates the "noise" of asynchronous communication and synchronization of parallel threads, on the other hand, lighter agents that perform simple but coarse learning techniques (like Random Walk) are being slowed down by the more computationally intensive learning algorithms (like Q-Learning). This may prove an interesting ground to cover in future experiments. The real-time competition between fast and simple learning strategies against slower but more refined ones can have interesting consequences in the effect of advice-exchange.

Although this was not an issue, the simulation runs faster than real-time, even when all agents are performing learning steps. Simulations ran (usually) for 1600 epochs, where each epoch consists of 50 green-yellow-red cycles, each consisting of 100 turns in which, on average, approximately 150 cars were moved and checked for collisions. Each simulation, with five disconnected crossings (i.e., four parallel learning algorithms and one heuristic agent), took 4 to 5 hours to run in a Pentium IV at 1.5 GHz. To generate a set of comparable data, this scenario must be run twice: with and without advice-exchange.

### 3.2 Learning Agents

This section describes the learning algorithms used by each of the agents involved in the experiments, as well as the heuristic used for the fixed strategy agent.

#### 3.2.1 Stand-alone agents

The stand-alone versions of the learning agents are used to provide results with which the performance of advice-exchanging agents could be compared. The stand-alone agents implement four classical learning algorithms: Random Walk (RL), which is a simple hill-climbing algorithm, Simulated Annealing (SA), Evolutionary Algorithms (EA) and Q-Learning (QL). A fifth agent was implemented (HEU) using a fixed heuristic policy. As the objective of these experiments was not to solve this problem in the most efficient way, but to evaluate advice-exchange for problems that have characteristics similar to this, the algorithms were not chosen or fine-tuned to produce the best possible results for traffic control. The choice of algorithms and their parameters was guided by the goal of comparing the performance of a heterogeneous group of learning agents, using classical learning strategies, in a non-deterministic, non-supervised, partially-observable problem, with and without advice-exchange.

All agents, except QL and HEU, adapt the weights of a small, one hidden layer, neural network. Experiments were conducted with several topologies, but the results discussed below refer to fully connected networks of 4x4x1, when using *count state representation*, and 8x4x1, when using *count-time state representation*. The weights of these networks were initialised randomly with values in the range [-0.5, 0.5]. The hidden layer is composed of sigmoids whose output varies in [-1, 1], while the outer layer sigmoids' output is in the range [0, 1]. This neural network will produce an output that will be the percentage of green-time (*gt*) for the next green-yellow-red cycle.

The Random Walk (RW) algorithm simply disturbs the current values of the weights of the neural network by adding a random value in the range [-*d, d*], where $d$ is the maximum disturbance, which will be updated after a given number of epochs, (*ne*), according to $d=\gamma d$, with $0<\gamma<1$, until it reaches a minimum value, (*min_d*). An epoch consists of $n$ green-red-yellow cycles. At the end of an epoch, the new set of parameters is kept if the average quality of service in the controlled crossing during that epoch is better than the best average quality achieved so far. The values used for the parameters of this algorithm in the experiments discussed here were in the following intervals: $d \in [0.5, 0.7]$, $min\_d=0.01$, $\gamma=0.99$, $n=50$, $n \in [3, 7]$. These values apply also for the decay of disturbance limits in the following descriptions of SA and EA. When referring to the intervals in which values were chosen, it is meant that in different experiments several combinations of parameter values were tested but the initial value for these parameters was always in the mentioned range.

Simulated Annealing (SA), (Kirkpatrick, Gelatt and Vecchi, 1983), works in a similar way to Random Walk, but it may accept the new parameters even if the quality has diminished. New parameters are accepted if a uniformly generated random number $p \in [0,1[$, is smaller than

$$pa(t) = e^{-\Delta q / T}, \quad (5)$$

where $T$ is a temperature parameter that is decreased during training in the same way as $d$ in RW and $\Delta q$ is the difference between the best average quality achieved so far and the average quality of the last epoch.

Evolutionary Algorithms (EA), (Holland, 1975; Koza, 92), were implemented in a similar way to the one described in (Glickman and Sycara, 1999), which is reported to have been successful in learning to navigate in a difficult variation of the maze problem by updating the weights of a small Recurrent Artificial Neural Network. This implementation relies almost totally in the mutation of the weights, in a way similar to the one used for the disturbance of weights described for RW and SA. Each set of parameters (specimen), which comprises all the weights of a neural network of the appropriate size for the state representation being used, is evaluated during one epoch. After the whole population is evaluated, the best $n$ specimens are chosen for mutation and recombination. An elitist strategy is used by keeping the best $b$ specimens untouched for the next generation. The remainder of the population is built as follows: the first $m$ are mutated, the remaining specimens ($r$) are created from pairs of the selected specimens, by choosing randomly from each of them entire layers of neural network weights. The values used for the parameters of this algorithm in the experiments discussed here were in the following intervals: $n \in [7, 10]$, $b \in [3, 7]$, $m \in [15, 25]$, $r \in [2, 5]$. The size of the population was [20, 30].

Q-Learning (QL), (Watkins and Dayan 1992), uses a lookup table with an entry for each state-action pair in which the expected utility $Q(s,a)$ is saved. $Q(s,a)$ represents the expected utility of doing action $a$ when the environment is in state $s$. Utility is updated in the usual way, i.e.,

$$Q(s,a) = Q(s,a) + \alpha(r + \beta Q \max(s') - Q(s,a)), \quad (6)$$

where $s'$ is the state after performing action $a$, $\alpha$ is the learning rate, $\beta$ the discount factor and $Q_{max}(s)$ is given by

$$Q_{\max}(s) = \max_{a}(Q(s,a)), \quad (7)$$

for all possible actions $a$ when the system is in state $s$.

The values of $\alpha$ (learning rate), in the different experiments, were in the interval [0.5,0.7]. The learning rate is updated after a given number of epochs, ($ne$), according to $\alpha = \gamma \alpha$, with $0 < \gamma < 1$, until it reaches a minimum value (which in this case was 0.012). In the experiments discussed here $ne=5$. Parameter $\beta$ (discount) was fixed in each experiment. Different values for $\beta$ were tested in several experiments within the interval [0.6,0.8]. The choice of action $a$, given that the system is in state $s$, was done with probability $p(a|s)$ that is given by a Boltzman distribution

$$p(a \mid s) = \frac{e^{Q(s,a)/T}}{\sum_{i} e^{Q(s,a_i)/T}}, \quad (8)$$

where $T$ is a temperature parameter whose initial value was in the interval [0.3,0.7] and was decayed in a similar way to the one described for $\alpha$.

Since the state of the environment is a real-valued vector, a partition of the space in a square lattice is required to map environment states (continuous) to internal (discrete) states. The decision of which is the state of the environment at a given time is made by calculating the Euclidean distance between the continuous valued world state and each of the discrete state representations and selecting the state with minimum distance. For the *count state representation* this partition consists in states composed of quadruples of the form: $(x_1, x_2, x_3, x_4)$, for which $x_1 + x_2 + x_3 + x_4 = 1.0$, and $x_i \in \{0, 0.1, 0.2, ... , 0.9, 1.0\}$. This reduction of the state space, compared to the use of all possible quadruples with elements in $\{0.0, 0.1, 0.2,...\}$, is possible given that the representation of the environment is composed of the percentages of vehicles in each lane relative to the number of vehicles in all lanes, thus being restricted to quadruples for which the sum of all elements is 1.0. For the *count-time state representation* the internal state is of the form: $(x_1, x_2, x_3, x_4, x_5, x_6, x_7, x_8)$, where the first four parameters are generated in the same fashion as in the previous case but with a coarser granularity and, the last four elements, are selected combinations of values in $\{0.0, 0.25, 0.5, 0.75, 1.0\}$. The number of states for the first and second case is, respectively, 286 and 1225. In future experiments, with more informative state-space representations, it may become necessary to use a neural network to map states to their correspondent utility as described in (Barto, Sutton and Watkins, 1990; Lin, 1992). Actions, i.e., green-time for the North and South lanes, are also considered as discrete values starting from zero, up to the maximum green time allowed, and differing by 0.05 steps.

The heuristic agent (HEU) gives a response that is calculated in different ways, depending on the state representation. The percentage of green-time ($g$) for the

sentation. The percentage of green-time (*g*) for the North and South lanes is calculated by

$$g = \frac{\max(n_N, n_S)}{\max(n_N, n_S) + \max(n_E, n_W)} \quad (9)$$

for the *count state representation*, and in a similar way accounting for the *lifetime* values for the first car in each track for the *count-time state representation*. The idea is that it seems reasonable to attribute a green time to the North and South lanes proportionally to the magnitude of the maximum number of cars (and the waiting times) relative to that of the sum of these values for both pairs of lanes.

### 3.2.2 Advice-exchange mechanism

The main expectation, when advice-exchange was chosen, was that using advice from the more knowledgeable agents in the system would improve the learning performances of all agents. Since supervision is a more efficient training method than reinforcement, (at the expense of needing more information) then, when no supervision information is available why not use advice as supervision? Better yet, if agents have different learning skills, which produce different types of progress through the search-space, they may be able to avoid that others get stuck in local minima by exchanging advice. It is unlikely that all agents are stuck in the same local minima and the exchange of information regarding the appropriate answers to some environment states could force others to seek better solutions.

The process of advice-exchange is conducted in a different way in the agents that use a neural network as activation function and in the Q-Learning agent. The heuristic agent does not participate in the experiments concerning advice-exchange. Advice-exchange is prohibited in the first 2 to 10 epochs of training, depending on the experiments, to avoid random advice being exchanged and to allow some time for the agents to announce a credible best average quality value. All agents broadcast their best result (i.e., best average quality measured during one epoch) at the beginning of each epoch.

At the beginning of each green-yellow-red cycle, agent *i* (the advisee) evaluates its current average quality ($cq_i$) since the beginning of the present epoch. This quality is compared with the best average quality ($bq_j$), for all agents *j*, broadcasted by other agents at the end of last epoch. Let $mbq_k = max(bq_j)$, for all agents $j \neq i$. If $cq_i < d\ mbq_k$ where *d* is a discount factor (usually 0.8), then agent *i* will request advice from agent *k* (the advisor) who as advertised the best average quality. The request for advice is sent having as parameter the current state of the environment as seen by agent *i*. The advisor switches his working parameters (neural network weights in most cases) to the set of parameters that was used in the epoch where the best average quality was achieved and runs the state communicated by the advisee producing its best guess at what would be the appropriate response to this state. This response (the advised percentage of green time for the north and south lanes) is communicated back to the advisee. In the case where advisees are RW, SA and EA agents, the communicated result is backpropagated as desired response, using the standard backpropagation rule (Rumelhart, Hinton and Wlliams, 1986) to update the weights of the network immediately after this pass. In some experiments an adaptive learning rate backpropagation (Silva and Almeida, 1990) was used but results were not significantly different. The values for the main backpropagation parameters used in the experiments discussed here were in the following intervals: learning rate: [0.001, 0.05], momentum: [0.3, 0.7].

Table 1: Steps of the advice-exchange sequence for an advisee agent (*i*) and an advisor agent (*k*).

| |
|---|
| 1. Agent *i*: receive the best average quality ($bq_j$) from all other agents ($j \neq i$). Quality for Agent *i* is $cq_i$. |
| 2. Agent *i*: get state *s* for evaluation. |
| 3. Agent *i*: calculate $k = \arg \max_j(bq_j)$, for all agents ($j \neq i$). |
| 4. Agent *i*: if $cq_i < d\ max(bq_j)$: <br>   a. Agent *i*: send agent *k* the current state *s* and request advice. <br>   b. Agent *k*: switch to best parameters and run state *s* to produce its best guess at the adequate response (*g*). <br>   c. Agent *k*: return *g* to Agent *i*. <br>   d. Agent *i*: process advice (*g*). |
| 5. Agent *i*: run state *s* and produce response *g'*. |

When the Q-Learning agent is the advisor, switching to best parameters corresponds simply in selecting the action with best quality. In the case where the Q-Learning agent is the advisee, the action that is closest to the given advice (recall that actions are discrete values in this case) is rewarded in a similar way to that described in (6). Since in this case the state of the system after action *a* is unknown, the value of $Q_{max}(s')$ is replaced by a weighted average of the utilities of all the possible following states when executing action *a* at state *s*:

$$Q_{max}(a) = p(s'|a,s) Q_{max}(s') \quad (10)$$

where $p(s'|a,s)$ is the probability of a transition to state s' given that action *a* is executed at state *s* and it is calculated based on previous experience, as the number of transitions ($nt_{s'as}$) to state *s'* when performing action *a* at the current state, *s*, relative to the total number of transitions from current state by action *a*, i.e.,

$$p(s'|a,s) = \frac{nt_{s'as}}{\sum_i nt_{ias}}, i \in S_{sa,} \quad (11)$$

where $S_{sa}$ is the set of states reachable from state *s* by action *a*. This type of adaptation of the state utility was proposed in Sutton's (1992) Dyna.

After updating the internal parameters with the advised information, the advisee agent gives the appropriate response to the system following the normal procedure for each particular algorithm.

## 4 Experimental Results

Before the discussion of the experimental results, let us put forward a few brief remarks concerning the simulation and experiments.

The type of problem we are dealing with is a difficult topic for simulation. Several works have been done in this area, and the simplifications made in this scenario were, in great measure, inspired by previous works mentioned in section 2. Nevertheless, the tuning of the simulator for the problem at hand was not a trivial matter. The problems tended to be either too easy or too hard, and, in the first experiments, only marginal differences could be observed in the quality measure during training. The most interesting experiments conducted were the cases where lanes had quite different behaviours from one another, ranging from a medium steady flow, to high peaks of traffic intermediated with periods with nearly no traffic at all. As observed in (Lin 1992), when doing similar experiments with variants of Q-Leaning, harder tests provided the best results for advice-exchange. However, there seems to be a fine line between hard solvable problems and, apparently, insoluble tasks in which no learning strategy, nor heuristic, could reach reasonable values of quality.

The interpretation of results is also not an easy task. The fact that agents are running online, and most of them are based on random disturbance, added to the stochastic nature of the environment, produces very "noisy" quality evaluations. The results presented here focus mainly on the analysis of the evolution of the best quality achieved up the present moment of training. Other measures also give us an insight on the process, but, at the present moment this seemed to be the one that could better illustrate the main observations made during experiments.

The above-mentioned stochastic nature of the problem, and the large simulation times, also forced a compromise in the choice of parameters for car generation. Although an even greater variety of behaviours could be achieved with other type of functions, whose periods span over a larger time-frame, this would require that each training epoch would be much longer, so that a comparison between values of different epochs would be fair. A lot of care was put into making epochs equally hard, in terms of frequency of cars generated.

One last remark concerning the discussion of results that will follow. The amount data necessary for a sound statistical comparison and evaluation of this technique is still being gathered. The preliminary results discussed here, produced in a series of 30 full trials, give us an insight on the problems and possible advantages of advice-exchange during learning, but data is still not sufficient for a detailed evaluation of the advantages and drawbacks of this technique. The above mentioned trials were ran under different conditions, either in the parameters of car-generation, lane-size and car speeds, or in the decay rates and other parameters of the algorithms themselves.

Before starting experiments, some results were expected, namely:

a) Initial disturbance of the learning process due to advice by non-expert peers, as reported by Tan (1993) for cooperation amongst Q-Learning agents.
b) After a few epochs, fast, step-like, increases in quality of response, as soon as one of the agents, found a better area of the state space and drove other agents that had poorer performances to that area.
c) Final convergence on better quality values than in tests where no advice is exchanged.
d) Problems of convergence when using excess of advice, or high learning rates when processing advice.
e) Improved resistance to bad parameterisation (special in algorithms like Simulated Annealing, which have parameters, like temperature, that are difficult to tune).

The actual observed results differed in some respects from expectations. The initial disturbance, or slower convergence, reported by Tan (1993) for Q-Learning agents, was not observed as a rule, although it occasionally happened. The exact opposite was also observed. In some experiments we can find agents that use advice climbing much faster to a reasonable quality plateau. Occasionally learning was much slower afterwards (probably a local maximum was reached) and this high initial quality value was gradually surpassed by the stand-alone algorithms during the rest of the training.

The second expectation, the appearance of high steps in the quality measure, due to advice from an agent that discovered a much better area of the search-space, was observed, but seems to be far less common than expected. Figure 3 shows a detail of the initial phase of a trial where we can see a typical situation of the described behaviour. The Simulated Annealing agent jumps to a high quality area, and "pulls" Random Walk and Q-Learning into that area in a few epochs. In this experiment the advice-exchanging algorithms did not stop at this quality plateau, being able to obtain better scores than their counterparts.

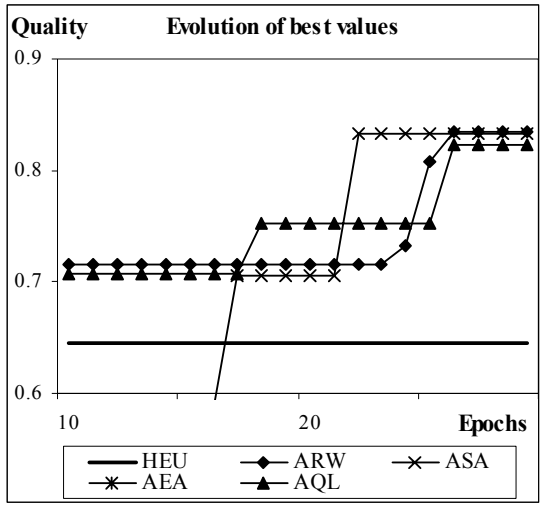

Figure 3: Detail of the initial phase of a trial where advice given by Simulated Annealing (ASA) led Random Walk (ARW) and Q-Learning (AQL) agents on a sudden climb of more than 10%. Evolutionary Algorithms also benefited from this jump, but the climb was less steep and from a lower point.

Results where the final quality values for the best agent, on trials with advice-exchange, is significantly better than in the normal case were observed, but do not seem to be as common as expected. Figures 4 to 7 show comparisons of the methods with and without advice-exchange for one of the trials where advice-exchange proved advantageous. Notice that all results are better than the one obtained by the heuristic agent (HEU), which was not frequent and denotes a particularly hard problem. The most usual result is that agents climb to the vicinity of the best agent's quality in few epochs, and make only minor improvements for rest of the trial.

The expectations referred in d) and e) were observed, as was foreseen. In fact, several cases were observed in trials without advice-exchange, where early freezing of the temperature parameter or the decay of the exploration rate, led to a sudden drop to a low-quality valley, from which the algorithm did not escape for the rest of trial. These events are rare in trials using advice-exchange.

One of the most interesting problems observed was that of ill advice. It was observed that some agents, due to a "lucky" initialisation and exploration sequence, never experience very heavy traffic conditions, thus, their best parameters are not suited to deal with this problem. When asked for advice regarding a heavy traffic situation, their advice is not only useless, but harmful, because it is stamped with the "quality" of an expert. In Q-Learning this was easy to observe because there were situations, far into the trials, for which advice was being given concerning states that had never been visited before. In the next section some measures to prevent this problem will be discussed.

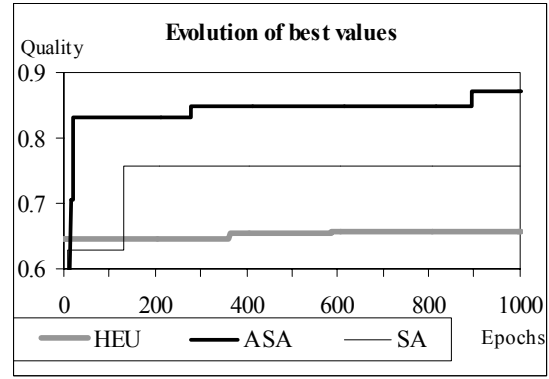

Figure 4: Comparison of Simulated Annealing performance, with (ASA) and without (SA) advice-exchange, and the corresponding heuristic (HEU) quality for the same trial.

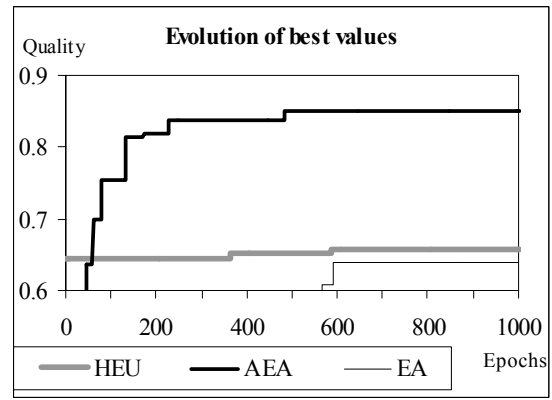

Figure 5: Comparison of Evolutionary Algorithms performance, with (AEA) and without (EA) advice-exchange, and the corresponding heuristic (HEU) quality for the same trial.

## 5 Conclusions and Future Work

As mentioned in the previous section, advice-exchange seems to be a promising way in which agents can profit from mutual interaction during the learning process. However, this is just the beginning of a search, where a few questions were answered and many were raised. A thorough analysis of the conditions in which this technique is advantageous is still necessary. It is important to discover how this technique performs when agents are not just communicating information about similar learning problems, but attempting to solve the same problem in a common environment. The application of similar methods to other type of learning agents, as well as other problems, is also an important step in the validation of this approach.

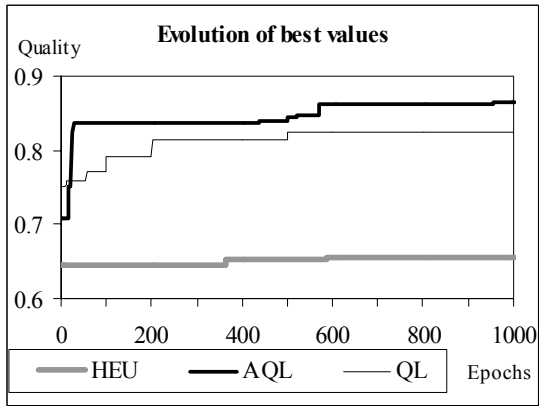

Figure 6: Comparison of Q-Learning performance, with (AQL) and without (QL) advice-exchange, and the corresponding heuristic (HEU) quality for the same trial

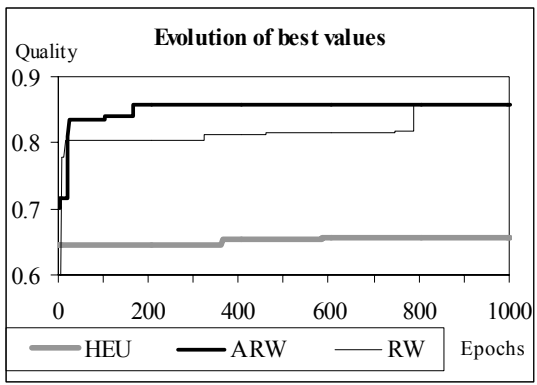

Figure 7: Comparison of Random Walk performance, with (ARW) and without (RW) advice-exchange, and the corresponding heuristic (HEU) quality for the same trial

For the time being, a more realistic traffic environment is under development based on the Nagel-Schreckenberg model for traffic simulation (Nagel and Shreckenberg, 1992). We hope that this new formulation provides a richer environment in which advice-exchange can be more thoroughly tested.

One of the main problems observed with advice-exchange is that bad advice, or blind reliance, can hinder the learning process, sometimes beyond recovery. One of the major hopes to deal with this problem is to develop a technique in which advisors can measure the quality of their own advice, and advisees can develop trust relationships, which would provide a way to filter bad advice. This may be especially interesting if trust can be associated with agent-situation pairs. This will allow the advisee to differentiate who is the expert on the particular situation it is facing. Work on "trust" has been reported recently in several publications, one of the most interesting for the related subject being (Sen, Biswas and Debnath, 2000).

Another interesting issue rises from the fact that humans usually offer unrequested advice for limit situations. Either great new discoveries or actions that may be harmful for the advisee seem to be of paramount importance in the use of advice. Rendering unrequested advice at critical points, by showing episodes of limit situations, also seems like a promising approach to improve the skills of a group of learning agents. The same applies to the combination of advice from several sources. These techniques may require an extra level of skills: more elaborate communication and planning capabilities, long-term memory, etc. These capabilities fall more into the realm of symbolic systems. The connection between symbolic and sub-symbolic layers, which has been also an interesting and rich topic of research in recent years, may play an important role in taking full advantage of some of the concepts outlined in this work. Our major aim is to, through a set of experiments, derive some principles and laws under which learning in the multi-agent system framework proves to be more effective, and inherently different from just having agents learning as individuals (even if they are together in the same environment).

## Acknowledgements

The authors would like to thank, Manuel Sequeira, Thibault Langlois, Jorge Louçã , Pedro Figueiredo, Ana Violante, Rui Lopes, Ricardo Ribeiro, Francisco Pires, Isabel Machado, Sofia Regojo and two anonymous reviewers. Also, our thanks to the ResearchIndex crew.

## References

C. Baroglio. Teaching by shaping. Proceedings ICML-95, Workshop on Learning by Induction vs. Learning by Demonstration, Tahoe City, CA, USA, 1995

A. G. Barto, R. S. Sutton and P. S. Brouwer. Associative search network: a reinforcement learning associative memory. Biological Cybernetics, 40(3):201-211, 1981

A. G. Barto, R. S. Sutton, C. J. C. H. Watkins. Learning and sequential decision making. Garb & J. W. Moore (Eds.), Learning and computational neuroscience. MIT Press, 1990

H. R. Berenji and D. Vengerov. Advantages of Cooperation Between Reinforcement Learning Agents in Difficult Stochastic Problems. 9th IEEE International Conference on Fuzzy Systems (FUZZ-IEEE '00), 2000

R. I. Brafman and M. Tennenholtz. On partially controlled multi-agent systems. Journal of Artificial Intelligence Research, 4:477-507, 1996

E. Brockfeld et al. Optimizing Traffic Lights in a Cellular Automaton Model for City Traffic. Physical Review E 64 , 2001

P. A. Castillo et al. SA-Prop: Optimization of Multilayer Perceptron Parameters using Simulated Annealing. IWANN99, 1998

J. A. Clouse. Learning from an automated training agent. Gerhard Weiß and Sandip Sen, editors, Adaptation and Learning in Multiagent Systems, Springer Verlag, Berlin, 1996

W. Ehardh et al. The Improvement and Comparison of different Algorithms for Optimizing Neural Networks on the MasPar {MP}-2. Neural Computation {NC}'98, ICSC Academic Press, Ed.M. Heiss, 617-623, 1998

M. Glickman and K. Sycara. Evolution of Goal-Directed Behavior Using Limited Information in a Complex Environment GECCO-99: Proceedings of the Genetic and Evolutionary Computation Conference, July 1999

C. Goldman and J. Rosenschein. Mutually supervised learning in multi-agent systems. Proceedings of the IJCAI-95 Workshop on Adaptation and Learning in Multi-Agent Systems, Montreal, CA., August 1995

J. H. Holland. Adaptation in Natural and Artificial Systems. University of Michigan Press, 1975

O. C. Jenkins, M. J. Matarić and S. Weber. Primitive-based movement classification for humanoid imitation. Proceedings, first IEEE-RAS international conference on humanoid robotics, Cambridge, MA, MIT, 2000

D. Kazakov and D. Kudenko. Machine Learning and Inductive Logic Programmimng for Multi-Agent Systems. Multi Agents Systems and Applications: 9th EACCAI advanced course, Selected Tutorial Papers, 246-271, Prague, Czech Republic, July 2001

S. Kirkpatrick, C. D. Gelatt and M. P. Vecchi. Optimization by simulated Annealing. Science, Vol. 220:671-680, May 1983

J. R. Koza. Genetic programming: On the Programming of Computers by Means of Natural Selection. MIT Press, Cambridge MA, 1992

K. W. C. Ku and M. W. Mak. Exploring the effects of Lamarckian and Baldwinian learning in evolving recurrent neural networks. Proceedings of the IEEE International Conference on Evolutionary Computation, 617-621, 1997.

L.-J. Lin. "Programming Robots Using Reinforcement Learning and Teaching. Proceedings of the American Association for Artificial Intelligence (AAAI-91), 781-786, 1991

L.-J. Lin. Self-improving reactive agents based on reinforcement learning, planning and teaching. Machine Learning 8:293-321, 1992

R. Maclin and J. Shavlik. Creating advicetaking reinforcement learners. Machine Learning 22:251-281, 1997

M. J. Matarić. Using Communication to Reduce Locality in Distributed Multi-agent learning. Brandeis University Computer Science Technical Report CS-96-190, 1996

M. J. Matarić. Learning in behaviour-based multi-robot sysytems: policies, models and other agents. Journal of Cognitive Systems Research 2:81-93, Elsvier, 2001

M. J. Matarić. Sensory-motor primitives as a basis for imitation: linking perception to action and biology to robotics. C. Nehaniv & K. Dautenhahn (Eds.), Imitation in animals and artifacts, MIT Press, 2001b

K. Nagel, M Shreckenberg. A Cellular Automaton Model for Freeway Traffic. J. Phisique I, 2(12):2221-2229, 1992

M. Nicoluescu and M. J. Matarić. Learning and interacting in human-robot domains. K. Dautenhahn (Ed.), IEEE Transactions on systems, Man Cybernetics, special issue on Socially Intelligent Agents – The Human In The Loop, 2001

E. Oliveira, J.M.Fonseca, N. Jennings. Learning to be competitive in the Market. AAAI'99 - American Association of Artificial Intelligence Workshop on Negotiation, Orlando, USA, 1999

D. E. Rumelhart, G. E. Hinton and R. J. Wlliams. Learning internal representations by error propagation. Parallel Distributed Processing: Exploration in the Microstructure of Cognition, vol. 1: Foundations, 318-362, Cambridge MA: MIT Press, 1986

R. Salustowicz. A Genetic Algorithm for the Topological Optimization of Neural Networks. PhD Thesis, Tech. Univ. Berlin, 1995

S. Sen. Reciprocity: a foundational principle for promoting cooperative behavior among self-interested agents. Proc. of the Second International Conference on Multiagent Systems, 322-329, AAAI Press, Menlo Park, CA, 1996

S. Sen, A. Biswas, S. Debnath. Believing others: Pros and Cons. Proceedings of the Fourth International Conference on Multiagent Systems, 279-286, 2000

F. M. Silva and L. B. Almeida. Speeding up back-propagation. Advanced Neural Computers, 151-158, A'dam North-Holland, 1990

R. S. Sutton and A. G. Barto A Temporal-Difference Model of Classical Conditioning. Tech Report GTE Labs. TR87-509.2, 1987

R. S. Sutton. Reinforcement learning architectures. Proceedings ISKIT'92 International Symposium on Neural Information Processing, Fukuoka, Japan, 1992.

M. Tan. Multi-Agent Reinforcement Learning: Independent vs. Cooperative Agents. Proceedings of the Tenth International Conference on Machine Learning, Amherst, MA, 330-337, 1993

T. Thorpe. Vehicle Traffic Light Control Using SARSA, Masters Thesis, Department of Computer Science, Colorado State University, 1997

A.P. Topchy, O.A. Lebedko and V.V. Miagkikh. Fast learning in multilayered neural networks by means of hybrid evolutionary and gradient algorithms Proc. of IC on Evolutionary Computation and Its Applications, Moscow, 1996

C. J. C H. Watkins and P. D. Dayan. Technical note: Q-learning. Machine Learning 8, 3:279-292, Kluwer Academic publishers, 1992

G. Weiß, P. Dillenbourg. What is 'multi' in multiagent learning. P. Dillenbourg (Ed.), Collaborative learning. Cognitive and computational approaches (Chapter 4, 64-80, Pergamon Press, 1999

S. D. Whitehead. A complexity Analisys of Cooperative Mechanisms in Reinforcement Learning. Proc. of the 9th National Conference on Artificial Inteligence (AAAI-91), 607-613, 1991

S. D. Whitehead and D. H. Ballard. A study of cooperative mechanisms for faster reinforcement learning. TR 365, Computer Science Department, University of Rochester, 1991

X. Yao. Evolving artificial neural networks. Proceedings of the IEEE, 87(9),1423-1447, 1999